\documentclass[letterpaper]{article} 
\usepackage{aaai2027} 
\usepackage[hyphens]{url} 
\usepackage{graphicx} 
\usepackage{natbib} 
\usepackage{caption} 
\usepackage{algorithm}
\usepackage{algorithmic}

\usepackage{amsmath}
\usepackage{amssymb}
\usepackage{booktabs}
\usepackage{multirow}
\usepackage{array}
\usepackage{pifont}
\usepackage{amsmath}
\usepackage{newfloat}
\usepackage{listings}
\DeclareCaptionStyle{ruled}{labelfont=normalfont,labelsep=colon,strut=off} 
\floatstyle{ruled}
\newfloat{listing}{tb}{lst}{}
\floatname{listing}{Listing}

\usepackage{booktabs}

\nocopyright

\title{Breaking Self-Attention Failure: Rethinking Query Initialization for Infrared Small Target Detection}

\author{
    Yuteng Liu\textsuperscript{\rm 1},
    Duanni Meng\textsuperscript{\rm 1},
    Yimian Dai\textsuperscript{\rm 2},
    Maoxun Yuan\textsuperscript{\rm 1},
    Xingxing Wei\textsuperscript{\rm 1},
    Bo li\textsuperscript{\rm 1}
}

\affiliations{
    \textsuperscript{\rm 1}School of Artificial Intelligence, Beihang University\\
    \textsuperscript{\rm 2}College of Computer Science, Nankai University\\
    liuyuteng@buaa.edu.cn, MengDuanni@buaa.edu.cn, yimian.dai@nankai.edu.cn, mxyuan@buaa.edu.cn, xingxingwei@buaa.edu.cn, boli@buaa.edu.cn
}

\begin{document}

\maketitle

\begin{abstract}
Infrared small target detection (IRSTD) faces significant challenges due to low signal-to-noise ratios, extremely small target sizes, and complex cluttered backgrounds. Although DETR-based detectors benefit from global context modeling, their query initialization can become unreliable in IRSTD because only a few encoder tokens correspond to targets, while the majority describe the background. We revisit this phenomenon and reveal that the target-relevant embeddings of IRST are inevitably overwhelmed by dominant background features due to the self-attention mechanism, leading to unreliable query initialization and inaccurate target localization. To address this issue, we propose SEF-DETR, a novel framework that refines query initialization through Patch-wise Spectral Screening (PSS), Frequency-Routed Examination (FRE), and Reliability-Consistency-aware Fusion (RCF). PSS encodes the radial and directional energy distributions of local Fourier spectra and aggregates overlapping patch predictions into a target-relevant density map. Guided by this map, FRE performs sparse deformable re-examination after each encoder layer, while retaining content-driven sampling to distinguish true targets from frequency-domain false alarms. RCF further re-ranks candidate queries according to spatial–frequency consistency and frequency reliability. Extensive experiments on three public IRSTD datasets demonstrate that SEF-DETR achieves superior detection performance over state-of-the-art methods with low computational overhead, providing a robust and efficient DETR-based solution for IRSTD.
\end{abstract}

\section{Introduction}
\label{sec:intro}

Infrared small target detection (IRSTD) is essential for a wide range of military and civilian applications, such as avian intrusion warning systems \cite{dai2021attentional}, maritime search and rescue \cite{yuan2022translation,yuan2024c2former}, and aerial surveillance \cite{zhao2025rethinking, yuan2025unirgb}. However, infrared small targets (IRST) are inherently difficult to identify due to their long imaging distances, lack of discriminative texture, and weak thermal contrast. They often manifest as faint, structureless blobs with extremely low signal-to-noise (SNR) and signal-to-clutter ratios (SCR) \cite{dai2023one,yuan2024improving}. Such characteristics make it challenging to distinguish targets from dynamic and cluttered backgrounds, especially when environmental noise dominates the thermal response. Consequently, designing a robust and efficient IRSTD method that can accurately localize small targets across complex infrared scenes remains a critical and unsolved problem.

\begin{figure}[t]
   \centering
   \includegraphics[width=\linewidth]{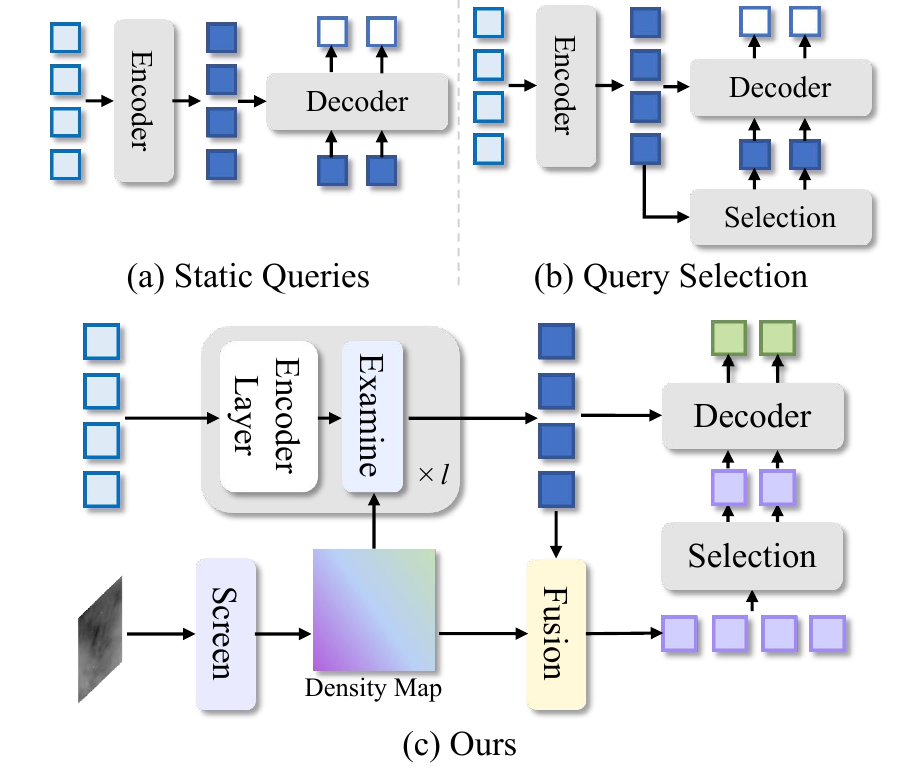}
   \caption{Comparison of different query initialization methods. \textbf{(a)} Static queries used for each inference. \textbf{(b)} Queries selected from the encoder output. \textbf{(c)} Our query initialization with screening, examination, and fusion mechanisms.}
   \label{fig:structure}
\end{figure}

Most current IRSTD methods are based on convolutional neural networks (CNNs), which preserve local spatial and texture cues. For example, DNA-Net \cite{li2022dense} mitigates deep-feature degradation caused by pooling, while MSHNet \cite{liu2024infrared} uses a lightweight multi-scale head for accurate localization. However, CNN have limited capacity to model the global context needed to distinguish dim targets from structured clutter. Recently, DETR \cite{carion2020end} reformulates detection as direct set prediction using Transformer encoder and decoder. Building on this, numerous DETR methods \cite{zhu2020deformable,liu2022dab,zhang2022dino,li2022dn} have been developed to enhance feature representation and accelerate convergence. However, these methods are inappropriate for IRSTD task because the IRST exhibit limited texture and occupy only a few pixels, making object query initialization susceptible to background noise rather than accurately locating the object. This motivates a task-specific question: \textit{how can we design an effective DETR-based detector specifically for IRSTD?}

\begin{figure}[t]
   \centering
\includegraphics[width=\linewidth]{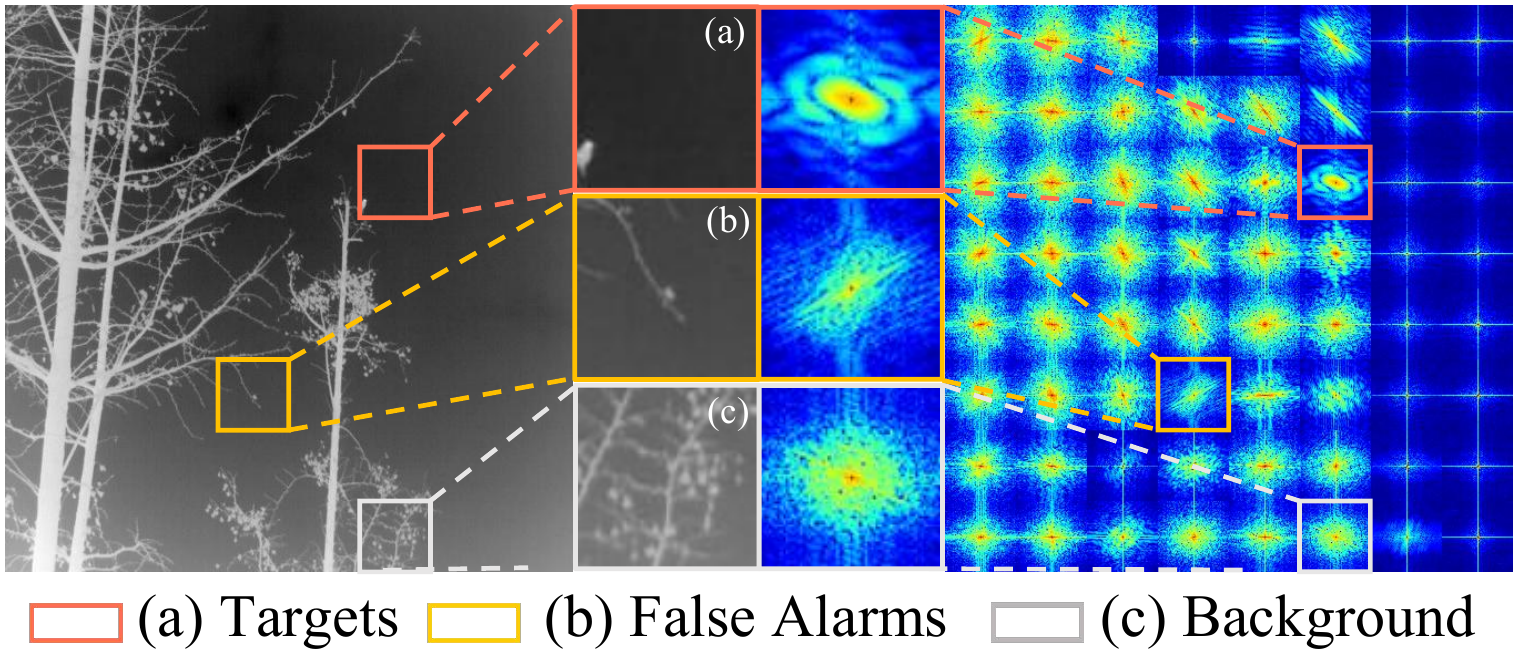}
   \caption{Illustration of local patches of IRST from the IRSTD-1k dataset. We compute the FFT spectrum for each local patch. The red, yellow, and gray boxes denote the target, target-like interference, and background, respectively.}
   \label{fig:frquency_map}
\end{figure}

To this end, we revisit self-attention from the \textit{embedding dilution} perspective (Sec. \ref{sec:analysis}) and analyze the possible reasons for such inadequacy of DETR. \textbf{Our analysis reveals that the target-relevant embeddings of infrared small targets are inevitably overwhelmed by dominant background features, leading to severely diluted representations due to the self-attention mechanism within DETR detectors.}
To address this issue, we observe that the Fourier spectrum of local patches provides a more discriminative cue for distinguishing small targets from both background clutter and target-like interference. As illustrated in Figure~\ref{fig:frquency_map}, patches containing true IRST (red box) exhibit frequency signatures that are markedly different from those of background regions (gray box) and distractors (yellow box).
Motivated by this insight, we utilize the Fourier spectrum of local patches to guide query initialization in DETR. Specifically, the frequency spectrum of each patch can be encoded into a frequency feature and fed into a classification head to determine target-relevant regions, where the corresponding embeddings will be prioritized. Thus, we propose a novel mechanism that sequentially performs patch \textbf{S}creening, encoder \textbf{E}xamination, and query \textbf{F}usion (\textbf{SEF}) for object query initialization (shown in Figure~\ref{fig:structure}(c)) in IRSTD task.

Specifically, we first propose a Patch-wise Spectral Screening (PSS) module that uses local Fourier spectra to generate a target-relevant density map, enabling the suppression of background-dominated embeddings. This density map is further exploited by a Frequency-Routed Examination (FRE) module to guide sparse, content-aware re-examination operations throughout the encoder layers. Finally, we design a Reliability-Consistency-aware Fusion (RCF) mechanism to refine query confidence by emphasizing regions where the spatial and frequency cues are both coherent and reliable, while suppressing inconsistent or uncertain responses. Building upon these components, we develop a new DETR-based architecture, termed \textbf{SEF-DETR}. This framework provides a principled solution to the embedding dilution issue and improves the performance of DETR-based object detectors in IRSTD task. In summary, our contributions in this paper are highlighted as follows: 


\begin{itemize}
    \item We revisit self-attention from the embedding dilution perspective and reveal that the target-relevant embeddings of IRST are inevitably overwhelmed by dominant background features due to the self-attention mechanism.
    \item We propose SEF-DETR, comprising Patch-wise Spectral Screening, Frequency-Routed Examination, and Reliability-Consistency-aware Fusion. This framework significantly improves the performance of DETR-based detectors on infrared small-target detection tasks while maintaining low computational overhead.
    \item Extensive experiments on three IRSTD datasets demonstrate that our SEF-DETR outperforms the previous state-of-the-art detectors and can be used as an effective DETR-based detector in the IRSTD task.
\end{itemize}

\section{Related Work}
\label{sec:relatedwork}
\subsection{DETR-based Object Detectors}
DETR \cite{carion2020end} revolutionized object detection with end-to-end set prediction. It eliminates hand-crafted components (anchors, NMS) by leveraging a transformer encoder-decoder and bipartite matching loss. However, its full-attention mechanism and under-optimized queries result in slow convergence and poor detection performance on specific tasks. To solve this problem, Deformable DETR \cite{zhu2020deformable} adopts sparse deformable sampling to improve efficiency. Besides, DAB-DETR \cite{liu2022dab} models queries as dynamic anchors for stable positional priors and DN-DETR \cite{li2022dn} uses de-noising training to simplify bipartite matching. Furthermore, DINO \cite{zhang2022dino} fuses these innovations to achieve superior performance across benchmarks. Additionally, RT-DETR \cite{zhao2024detrs} achieves real-time detection speed by simplifying the encoder.
However, above DETR-based methods are mainly designed for general object detection and cannot achieve superior performance on the IRSTD task. To this end, we delved into why DETR-based detectors struggle to handle IRSTD and propose a novel framework called SEF-DETR.

\subsection{Transformer-based IRSTD Methods}
Infrared small target detection remains challenging due to low contrast, small target size, and complex background clutter. Recently, Transformer-based methods have been widely adopted for their strengths in global contextual modeling and long-range feature interactions. TCI-Former~\cite{chen2024tci} draws inspiration from thermal conduction theory, introducing a pixel movement differential equation to refine target regions progressively. HSTNet \cite{li2025hstnet} proposes a hybrid spatial-channel sparse Transformer with dilated attention to maintain details while capturing dependencies. SCTransNet \cite{yuan2024sctransnet} designs cross Transformer blocks to mitigate semantic gaps in U-shaped networks, and IR-TransDet~\cite{lin2023ir} leverages a dual-branch CNN-Transformer structure to enhance robustness in low signal-to-noise scenarios. Furthermore, ISTD-DETR \cite{yang2025istd} integrates super-resolution preprocessing and state space modules into an enhanced RT-DETR framework. 

While these methods achieve superior performance by introducing transformer structure, they fail to diagnose the fundamental limitations of transformer in IRSTD task. Instead, we identify that target-relevant embeddings suffer from dilution during self-attention, and accordingly propose a novel framework to enhance target-relevant embeddings representations, thereby further improving IRSTD performance.


\section{Method}

\begin{figure}[t]
   \centering
\includegraphics[width=0.8\linewidth]{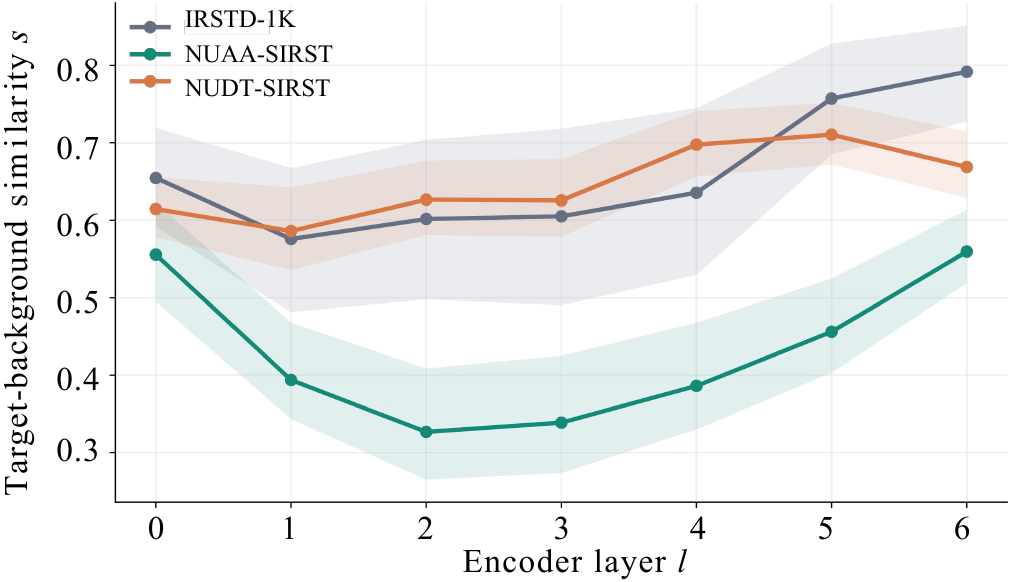}
   \caption{Target--background embedding similarity \(s\) from encoder layers \(l=0\) to \(6\) on IRSTD-1k, NUAA-SIRST, and NUDT-SIRST. The \(s\) increases with depth across all datasets.}
   \label{fig:sim_curve}
\end{figure}

\subsection{Analysis}
\label{sec:analysis}

\noindent \ding{202}~\textbf{Revisiting self-attention in DETR.}
Given an infrared feature map $X \in \mathbb{R}^{H \times W \times C}$, where $H$ and $W$ are its spatial dimensions and $C$ is its channel dimension, we flatten it into a sequence $\{x_i \mid i = 1, 2, \ldots, N\}$ with $N=HW$. A learnable embedding function $\mathcal{F}(\cdot)$ then produces
\begin{equation}
    y = [\mathcal{F}(x_1), \mathcal{F}(x_2), \ldots, \mathcal{F}(x_N)] + \mathcal{P},
\end{equation}
where $\mathcal{P}$ denotes learnable positional embeddings that encode spatial priors.
Within the DETR framework, object queries attend to the encoded feature embeddings through a multi-head attention mechanism, enabling global context aggregation. Specifically, the attention weights between query $i$ and key $j$ are computed as:
\begin{equation}
    A_{ij} = \frac{(y_i W^Q)(y_j W^K)^{\top}}{\sqrt{D}},
\end{equation}
and the output embedding  for each query is formulated as:
\begin{equation}
    z_i = \sum_j \sigma(A_{ij}) \, y_j W^V,
    \label{Equ:attn}
\end{equation}
where $W^Q, W^K, W^V \in \mathbb{R}^{D \times D}$ are learnable projection matrices, and $\sigma(\cdot)$ denotes the softmax normalization.

\begin{figure*}[!t]
   \centering
   \includegraphics[width=1.0\textwidth]{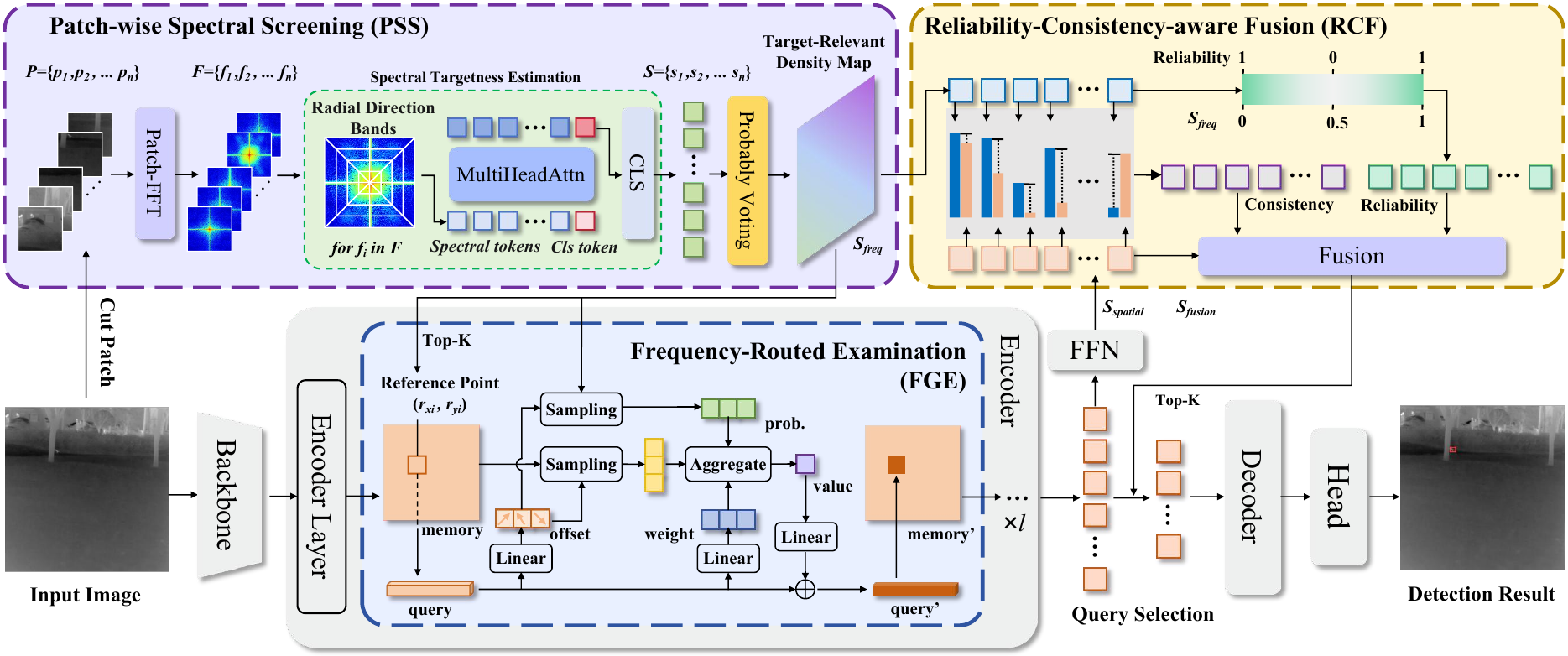}
   \caption{Overview of SEF-DETR. PSS produces a pixel-wise target-relevant density map, FRE routes sparse re-examination through the encoder, and RCF uses spatial--frequency reliability and consistency to select decoder queries.}
   \label{fig:overview}
\end{figure*}

\noindent\ding{203}~\textbf{Analysis from the embedding-dilution perspective.}
In the infrared small target detection task, only a few embeddings correspond to the target regions, while the vast majority of embeddings originate from background areas. Denoting $\Omega_t$ as the set of indices belonging to the target region and $\Omega_b$ as those of the background ($\Omega_t \cup \Omega_b = {1,\dots,N}$, $|\Omega_t| \ll |\Omega_b|$), hence, Equation (\ref{Equ:attn}) can be decomposed as:

\begin{equation}
    z_i = \underbrace{\sum_{j \in \Omega_t} \sigma(A_{ij}) y_j W^V}_{\text{target contribution}} \;+\;
           \underbrace{\sum_{j \in \Omega_b} \sigma(A_{ij}) y_j W^V}_{\text{background contribution}}.
    \label{Eq:attn_depose}
\end{equation}

Token imbalance alone does not prove that the background term dominates, because attention can in principle concentrate on the few target tokens. The failure arises when self-attention does not assign those tokens a sufficiently disproportionate mass. Define
\begin{equation}
\alpha_t^{(i)}=\sum_{j\in\Omega_t}\sigma(A_{ij}), \qquad
\alpha_b^{(i)}=\sum_{j\in\Omega_b}\sigma(A_{ij})=1-\alpha_t^{(i)}.
\end{equation}
Because $|\Omega_t|\ll|\Omega_b|$, preserving a target requires a much larger average weight per target token than per background token. When this concentration does not emerge, $\alpha_b^{(i)}$ dominates and repeated mixing makes target representations progressively resemble the background. We therefore treat embedding dilution as an empirically testable failure mode, not as a consequence of normalization alone. To measure it, we define the encoded feature set after the $l$-th layer as:
\begin{equation}
\mathbf{P}^{(l)} = [p^{(l)}_1, p^{(l)}_2, \ldots, p^{(l)}_N], \quad p^{(l)}_j \in \mathbb{R}^D.
\label{eq:encoded_features}
\end{equation}
We then compute the mean embeddings for the target and background regions:
\begin{equation}
\bar{p}^{(l)}_t = \frac{1}{|\Omega_t|} \sum_{m \in \Omega_t} p^{(l)}_m, \quad
\bar{p}^{(l)}_b = \frac{1}{|\Omega_b|} \sum_{n \in \Omega_b} p^{(l)}_n.
\end{equation}
To quantify the mixing between target and background features, we compute the cosine similarity between these two mean embeddings as:
\begin{equation}
c^{(l)} = \frac{|\bar{p}^{(l)T}_t \bar{p}^{(l)}_b|}{\|\bar{p}^{(l)}_t\|_2 \|\bar{p}^{(l)}_b\|_2}
\label{Eq:cross}
\end{equation}
which measures the resemblance between the aggregated target and background representations. We average it over the \(M\) images as \(s=\frac{1}{M}\sum_{t=1}^{M} c_t^{(l)}\). Figure~\ref{fig:sim_curve} reports \(s\) from layers \(l=0\) to \(l=6\) on all three datasets. Its consistent increase shows that target and background representations gradually lose distinctiveness within the self-attention in DETR. This is detrimental to preserving target-relevant embeddings, which validates our analysis.



\subsection{SEF-DETR}

To address the embedding-dilution issue in DETR, we propose SEF-DETR, a framework that sequentially performs patch Screening, encoder \textbf{E}xamination, and query \textbf{F}usion. The overall pipeline is illustrated in Figure~\ref{fig:overview}. It consists of \ding{202} Patch-wise Spectral Screening (PSS), \ding{203} Frequency-Routed Examination (FRE), and \ding{204} Reliability-Consistency-aware Fusion (RCF). 

\noindent \ding{202} \textbf{Patch-wise Spectral Screening.}
PSS exploits the observation that a compact infrared target and structured clutter may look similar in intensity but distribute their local energy differently across frequency scale and direction. Given an input image \(I\in\mathbb{R}^{H\times W}\), we extract overlapping patches \(\mathcal{P}=\{P_j\}_{j=1}^{J}\) with window size \(p\times p\) and stride \(s=p/2\), and compute the spectrum
\begin{equation}
\mathcal{F}_j=\operatorname{FFT2}(P_j), \qquad A_j=|\mathcal{F}_j|.
\end{equation}
We partition \(A_j\) into four radial ranges and eight angular sectors. Their Cartesian product yields 32 radial--directional bands \(\{B_{r,d}\}_{r=1,d=1}^{4,8}\), where radius represents frequency scale and angle represents orientation. Each band is encoded as a spectral token
\begin{equation}
\mathbf{t}_{j,r,d}=\phi(A_j\odot B_{r,d})
    +\mathbf{e}^{\mathrm{rad}}_{r}+\mathbf{e}^{\mathrm{ang}}_{d},
\label{Eq:pss_token}
\end{equation}
where \(\phi(\cdot)\) is a lightweight band encoder, while \(\mathbf{e}^{\mathrm{rad}}_{r}\) and \(\mathbf{e}^{\mathrm{ang}}_{d}\) preserve the physical meaning of each token. A learnable classification token interacts with all 32 spectral tokens through a multi-head self-attention encoder. Its output summarizes which scales and directions jointly support a compact target, and a classification head predicts the patch score
\begin{equation}
\mathbf{X}_j=\mathcal{T}\!\left(
[\mathbf{t}_{\mathrm{cls}};\{\mathbf{t}_{j,r,d}\}_{r,d}]
\right), \qquad
s_j=\sigma(\mathbf{w}_{c}^{\top}\mathbf{X}_j[0]+b_c).
\label{Eq:s_j}
\end{equation}



Overlapping patch predictions are finally assembled by \emph{target-relevant Hough voting}. Let \(K_j(x,y)\) be a normalized spatial vote cast by patch \(P_j\) over its support. The target-relevant density map is
\begin{equation}
S_{\mathrm{freq}}(x,y)=
\frac{\sum_{j=1}^{J}s_j K_j(x,y)}
{\sum_{j=1}^{J}K_j(x,y)+\epsilon}.
\label{Eq:hough}
\end{equation}
A true target is covered by several neighboring patches, so their votes reinforce one spatial hypothesis, while an isolated clutter response receives less coherent support. The resulting \(S_{\mathrm{freq}}\in[0,1]^{H\times W}\) is therefore an evidence map rather than a final detection: it indicates where the spatial encoder should spend additional modeling capacity.

\noindent \ding{203} \textbf{Frequency-Routed Examination.}
Embedding dilution develops progressively as encoder layers repeatedly mix a few target tokens with abundant background tokens. FRE is consequently inserted after every encoder layer rather than applied once after the encoder. Let \(\overline{\mathbf{M}}^{(\ell)}\) denote the multi-scale memory produced by the \(\ell\)-th standard encoder layer. FRE selects the top-\(K\) reference locations \(\mathcal{R}^{(\ell)}=\operatorname{TopK}(S_{\mathrm{freq}},K)\), maps them to the corresponding feature levels, and samples review queries \(\mathbf{q}^{(\ell)}_k=\overline{\mathbf{M}}^{(\ell)}(\mathbf{r}^{(\ell)}_k)\).
Each review query predicts deformable offsets and content-attention logits,
\begin{equation}
\Delta\mathbf{p}^{(\ell)}_k=W_{\Delta}^{(\ell)}\mathbf{q}^{(\ell)}_k,
\qquad
\mathbf{a}^{(\ell)}_k=W_{a}^{(\ell)}\mathbf{q}^{(\ell)}_k.
\label{Eq:fre_predict}
\end{equation}
The offsets remain query-driven: the frequency map chooses \emph{where to review}, while the current spatial feature decides \emph{where to look around that candidate}. At each predicted sampling location \(\mathbf{p}^{(\ell)}_{khmn}\), we sample both a memory value \(\mathbf{v}^{(\ell)}_{khmn}\) and frequency evidence \(e^{(\ell)}_{khmn}=S_{\mathrm{freq}}(\mathbf{p}^{(\ell)}_{khmn})\). Their calibrated attention weight is
\begin{equation}
\begin{aligned}
\alpha^{(\ell)}_{khmn}
&=\operatorname{softmax}_{m,n}\!\left(
a^{(\ell)}_{khmn}
+\lambda_{\ell}\log(e^{(\ell)}_{khmn}+\epsilon)
\right),\\
\lambda_{\ell}&=\operatorname{softplus}(\eta_{\ell}),
\end{aligned}
\label{Eq:fre_weight}
\end{equation}
where \(h\), \(m\), and \(n\) index attention head, feature level, and sampling point. The first term asks whether sampled content matches the review query, the second asks whether that location is also supported by spectral evidence. The independently learnable \(\lambda_{\ell}\) allows weak guidance in shallow layers, whose features are not yet fully discriminative, and stronger or weaker guidance at later depths according to training.

The sampled values are aggregated as
\begin{equation}
\mathbf{o}^{(\ell)}_k=W_o^{(\ell)}
\operatorname*{Concat}_{h}
\left[
\sum_{m,n}\alpha^{(\ell)}_{khmn}
\mathbf{v}^{(\ell)}_{khmn}
\right],
\label{Eq:fre_aggregate}
\end{equation}
and written back to the memory at \(\mathbf{r}^{(\ell)}_k\). Hence FRE does not force all high PSS responses to become stronger. It allocates an additional, sparse deformable examination to them, allowing feature interaction to recover missed targets while rejecting frequency-domain false alarms.

\noindent \ding{204} \textbf{Reliability-Consistency-aware Fusion.}
RCF integrates spatial and frequency evidence to rank target-relevant queries for the Transformer decoder. Let $S_{spatial} \in \mathbb{R}^{H_f \times W_f}$ be the sigmoid-normalized spatial confidence map obtained from the examined encoder features. Each candidate location $(u,v)$ has two normalized scores, $S_{spatial}(u,v)$ and $S_{freq}(u,v)$. We define two quantities:
\begin{itemize}
    \item Consistency $\mathbf{C}$ measures the agreement between the spatial and frequency domains:
    \begin{equation}
    C = 1 - | S_{spatial}(u,v) - S_{freq}(u,v) |.
    \end{equation}
    \item Reliability $\mathbf R$ quantifies the confidence of the frequency prior itself, being highest when the score is near 0 or 1:
    \begin{equation}
    R = 2 \cdot | S_{freq}(u,v) - 0.5 |.
    \end{equation}
\end{itemize}
Finally, the confidence score $S_{final}$ for each query is calculated using the following fusion function:
\begin{equation}
S_{final}(u,v) = S_{spatial}(u,v) \cdot \left( 1 + C \cdot (1 + R) \right).
\end{equation}
This formulation retains $S_{spatial}$ as the primary detection cue, while the $(C \cdot (1 + R))$ term adaptively amplifies scores when both domains are reliable and consistent. Therefore, the top-$K$ locations with the highest $S_{final}$ score are selected as the target-relevant queries for the decoder. 
By coupling the RCF module with the PSS and FRE, the frequency prior is converted into reliable spatial-frequency consensus for query selection, thereby preserving target-relevant queries while suppressing background-induced false positives.

\subsection{Loss Function}
To supervise target-relevant density map obtained by the PSS module, we introduce the patch-level classification loss $\mathcal{L}_{\mathrm{PSS}}$ during end-to-end training.
Given the ground-truth target set $\mathcal{G}$, a patch $P_j$ is assigned a positive label if it contains the center $\mathbf{c}_g$ of at least one target:
\begin{equation}
y_j=\mathbb{I}\!\left[\exists\,g\in\mathcal{G}:\mathbf{c}_g\in P_j\right].
\end{equation}
Since the number of background patches is substantially larger than that of target-containing patches, we employ a class-balanced focal loss to reduce the contribution of abundant easy negatives:
\begin{equation}
\begin{aligned}
\mathcal{L}_{\mathrm{PSS}}
=-\frac{1}{J}\sum_{j=1}^{J}\Big[
&\alpha y_j(1-s_j)^{\gamma}\log s_j \\
&+(1-\alpha)(1-y_j)s_j^{\gamma}\log(1-s_j)
\Big],
\end{aligned}
\label{Eq:pss_loss}
\end{equation}
where $s_j$ is the target-presence score predicted for patch $P_j$, $\alpha$ balances positive and negative samples, and $\gamma$ down-weights well-classified patches so that PSS focuses on ambiguous targets and target-like background regions.
Therefore, the overall training objective is stated as follows:
\begin{equation}
\mathcal{L}
=
\mathcal{L}_{\mathrm{Hungarian}}
+
\lambda\mathcal{L}_{\mathrm{PSS}},
\label{Eq:overall_loss}
\end{equation}
where $\mathcal{L}_{hungarian}$ is the Hungarian loss designed in DETR~\cite{carion2020end}, which consists of $L_1$ loss, GIoU loss and focal loss~\cite{lin2017focal}.
The hyperparameter $\lambda=2$ is used to control the balance between the two loss.


\section{Experiment}

\begin{table*}[t]
\small
\centering
\setlength{\tabcolsep}{0.9mm}
\renewcommand{\arraystretch}{0.9}
\begin{tabular}{@{}l|c|ccc|ccc|cc@{}}
\toprule
\textbf{Method} &
\textbf{Type} &
\textbf{P} &
\textbf{R} &
\textbf{F1} &
\textbf{AP} &
\textbf{AP\textsubscript{50}} &
\textbf{AP\textsubscript{vt}} &
\textbf{Params (M)} &
\textbf{GFLOPs} \\
\midrule

EFLNet \cite{yang2024eflnet}
& \multirow{4}{*}{CNN-based}
& 87.0 & 81.7 & 84.3
& 36.6 & 83.7 & 29.9 & 38.34 & 65.93 \\

YOLOv8m
&
& 88.0 & 78.5 & 83.0
& 34.5 & 79.2 & 28.3 & 25.90 & 50.77 \\

PConv \cite{yang2025pinwheel}
&
& 84.8 & 82.3 & 83.5
& \underline{38.3} & 79.8 & 29.0 & 28.83 & 58.85 \\

NS-FPN \cite{yuan2025ns}
&
& 87.7 & 80.1 & 83.7
& 37.4 & 81.2 & \underline{30.2} & 27.96 & 60.78 \\
\midrule

Deform-DETR \cite{zhu2020deformable}
& \multirow{7}{*}{DETR-like}
& 79.5 & 75.6 & 77.5
& 31.1 & 76.1 & 23.9 & 40.69 & 56.64 \\

DAB-DETR \cite{liu2022dab}
&
& 85.2 & 73.2 & 78.7
& 34.1 & 78.2 & 28.4 & 46.54 & 71.20 \\

DN-DETR \cite{li2022dn}
&
& 83.0 & 75.2 & 78.9
& 34.2 & 77.3 & 27.5 & 46.54 & 71.20 \\

RT-DETR \cite{zhao2024detrs}
&
& 86.2 & 79.5 & 82.7
& 37.2 & 81.8 & 28.2 & 42.73 & 44.90 \\

LW-DETR \cite{chen2024lw}
&
& 85.2 & 79.2 & 82.1
& 33.6 & 78.5 & 26.5 & 40.12 & 33.55 \\

RF-DETR \cite{robinson2026rfdetr}
&
& 84.5 & 82.6 & 83.6
& 36.3 & 80.3 & 27.4 & 33.37 & 64.73 \\

DINO \cite{zhang2022dino}
&
& 86.2 & \underline{85.4} & \underline{85.8}
& 37.1 & \underline{84.5} & 29.6 & 45.14 & 80.94 \\
\midrule

RT-DETR + SEF
& \multirow{3}{*}{SEF-based}
& \underline{88.8}$_{(\uparrow2.6)}$ & 79.9$_{(\uparrow0.4)}$ & 84.1$_{(\uparrow1.4)}$
& 38.4$_{(\uparrow1.2)}$ & 82.7$_{(\uparrow0.9)}$ & 29.7$_{(\uparrow1.5)}$ & 43.04$_{(+0.3)}$ & 45.08$_{(+0.2)}$ \\

LW-DETR + SEF
&
& 86.4$_{(\uparrow1.2)}$ & 81.3$_{(\uparrow2.1)}$ & 83.8$_{(\uparrow1.7)}$
& 36.5$_{(\uparrow2.9)}$ & 80.4$_{(\uparrow1.9)}$ & 28.6$_{(\uparrow2.1)}$ & 40.43$_{(+0.3)}$ & 33.73$_{(+0.2)}$ \\

\textbf{SEF-DETR}
&
& \textbf{92.4$_{(\uparrow6.2)}$} & \textbf{85.9$_{(\uparrow0.5)}$} & \textbf{89.0$_{(\uparrow3.2)}$}
& \textbf{38.9$_{(\uparrow1.8)}$} & \textbf{86.7$_{(\uparrow2.2)}$} & \textbf{32.8$_{(\uparrow3.2)}$}
& 46.70$_{(+1.6)}$ & 81.16$_{(+0.2)}$ \\
\bottomrule
\end{tabular}
\caption{Comparison with detection-based methods on IRSTD-1k. Both CNN-based and DETR-like detectors output bounding boxes with instance-level confidence scores, reporting P/R/F1 and AP metrics under the same detection setting.}
\label{tab:detection_compare}
\end{table*}

\begin{table*}[t]
\small
\centering
\setlength{\tabcolsep}{4mm}
\renewcommand{\arraystretch}{1.0}
\begin{tabular}{@{}l|ccc|ccc|ccc@{}}
\toprule
\multirow{2}{*}{\textbf{Method}} &
\multicolumn{3}{c|}{\textbf{IRSTD-1k}} &
\multicolumn{3}{c|}{\textbf{NUAA-SIRST}} &
\multicolumn{3}{c}{\textbf{NUDT-SIRST}} \\
&
\textbf{P} & \textbf{R} & \textbf{F1} &
\textbf{P} & \textbf{R} & \textbf{F1} &
\textbf{P} & \textbf{R} & \textbf{F1} \\
\midrule
MDvsFA \cite{wang2019miss}
& 55.0 & 48.3 & 47.5
& 84.5 & 50.7 & 59.7
& 60.8 & 19.2 & 26.2 \\

AGPCNet \cite{zhang2021agpcnet}
& 41.5 & 47.0 & 44.1
& 39.0 & 81.0 & 52.7
& 36.8 & 68.4 & 47.9 \\

ACMNet \cite{dai2021asymmetric}
& 67.9 & 60.5 & 64.0
& 76.5 & 76.2 & 76.3
& 73.2 & 74.5 & 73.8 \\

ISNet \cite{zhang2022isnet}
& 71.8 & 74.1 & 72.9
& 82.0 & 84.7 & 83.4
& 74.2 & 83.4 & 78.5 \\

ACLNet \cite{dai2021attentional}
& 84.3 & 65.6 & 73.8
& 84.8 & 78.0 & 81.3
& 86.8 & 77.2 & 81.7 \\

DNANet \cite{li2022dense}
& 76.8 & 72.1 & 74.4
& 84.7 & 83.6 & 84.1
& 91.4 & 88.9 & 90.1 \\

HCFNet \cite{xu2024hcf}
& 76.7 & \underline{74.3} & 75.5
& 85.5 & \underline{93.8} & 89.5
& 93.0 & 91.1 & 92.0 \\

IRSAM \cite{zhang2024irsam}
& 85.6 & 72.7 & 78.6
& \underline{92.7} & 89.4 & 91.0
& 93.6 & \underline{92.9} & 93.2 \\

MSHNet \cite{liu2024infrared}
& 82.1 & 72.4 & 76.9
& 91.0 & 89.4 & 90.2
& 89.2 & 88.6 & 88.9 \\

SCTransNet \cite{yuan2024sctransnet}
& 86.9 & 72.4 & 79.0
& 88.1 & 92.0 & 90.0
& 93.5 & 91.6 & 92.5 \\

MPCNet \cite{zhang2026mpcnet}
& \underline{90.3} & 70.4 & \underline{79.1}
& 90.8 & 93.6 & \underline{92.2}
& \underline{95.1} & 92.6 & \underline{93.9} \\

\textbf{SEF-DETR (Ours)}
& \textbf{92.4} & \textbf{85.9} & \textbf{89.0}
& \textbf{94.8} & \textbf{97.3} & \textbf{96.1}
& \textbf{100.0} & \textbf{96.3} & \textbf{98.1} \\
\bottomrule
\end{tabular}
\caption{Comparison with segmentation-based IRSTD methods on IRSTD-1k, NUAA-SIRST, and NUDT-SIRST. Their predicted masks are converted into bounding boxes using connected components. Since these boxes do not carry instance-level confidence scores, only P/R/F1 are reported. Best results are highlighted in \textbf{bold} and the second-place results are highlighted in \underline{underline}.}
\label{tab:segmentation_compare}
\end{table*}

\subsection{Datasets and Evaluation Metrics}

\noindent \textbf{Datasets.} We conduct comprehensive evaluations on three publicly available infrared small target detection benchmarks: IRSTD-1k \cite{zhang2022isnet}, NUAA-SIRST \cite{dai2021asymmetric}, and NUDT-SIRST \cite{li2022dense}. These datasets provide both bounding box annotations and pixel-wise segmentation masks, supporting evaluation under detection and segmentation paradigms. NUAA-SIRST contains 427 images, NUDT-SIRST contains 1,327 images, and IRSTD-1k contains 1,000 challenging images with precise annotations. For every compared method, we use the same fixed training, validation, and test partitions with a 3:1:1 ratio, following the prior IRSTD evaluation protocol and keeping target distributions balanced across splits.

\noindent \textbf{Metrics.} We evaluate all methods under a unified bounding-box detection setting. For segmentation methods, each connected component in the predicted mask is converted into a box using its horizontal bounding rectangle. Since these boxes without comparable instance-level confidence scores, they cannot be used to construct threshold-swept AP curves. Therefore, we report Precision, Recall, and F1 for these methods. Detection methods directly output boxes with confidence scores and are additionally evaluated using AP metrics. Following the AI-TOD tiny-object protocol \cite{wang2021tiny}, AP is averaged over box-IoU thresholds from 0.50 to 0.95 in steps of 0.05, AP\textsubscript{50} uses an IoU threshold of 0.50, and AP\textsubscript{vt} evaluates targets with an area below \(8^2\) pixels. Pixel-level IoU are not reported because SEF-DETR focuses on bounding-box detection rather than mask segmentation.

\subsection{Implementation Details}

SEF-DETR is built on DINO \cite{zhang2022dino} with a ResNet-50 backbone. We train for 120 epochs with batch size 2 on an NVIDIA GeForce RTX 4090, using DINO's random-crop and scale augmentations. AdamW \cite{loshchilov2017decoupled} uses an initial learning rate of \(10^{-4}\), followed by a tenfold decay. All experiments are conducted with three random seeds (42/43/44) and the averaged results are reported. PSS uses patch size \(p=64\), stride \(s=32\), 4 radial ranges, 8 angular sectors, a 64-dimensional spectral-token encoder with 4 attention heads. FRE uses 8 review anchors per feature level after each encoder layers, with an independent learnable evidence strength \(\lambda_{\ell}\) in every layer. RCF retains the top 300 queries. All other hyperparameters follow the default settings of DINO.

\subsection{Comparison with State-of-the-Art Methods}


We first compare SEF-DETR with CNN and DETR-based detectors on IRSTD-1k under the same bounding-box detection setting, using P/R/F1 and AP-based metrics. As shown in Table~\ref{tab:detection_compare}, SEF-DETR achieves the best overall performance, with 38.9\% AP, 86.7\% AP\textsubscript{50}, 32.8\% AP\textsubscript{vt}, 92.4\% precision, 85.9\% recall, and 89.0\% F1. The largest gain occurs on very tiny targets, where background-dominated query initialization is most damaging, supporting the effectiveness of SEF-DETR in mitigating the problem of embedding-dilution. We further integrate the proposed modules into RT-DETR and LW-DETR, yielding AP gains of 1.2 and 2.9 points, while introducing only 0.31M parameters and 0.18 GFLOPs to each detector, demonstrating that the proposed design is transferable across different DETR architectures. Additional results on other datasets are provided in the \textit{supplementary material}.

\begin{figure*}[t]
   \centering
   \includegraphics[width=0.95\textwidth]{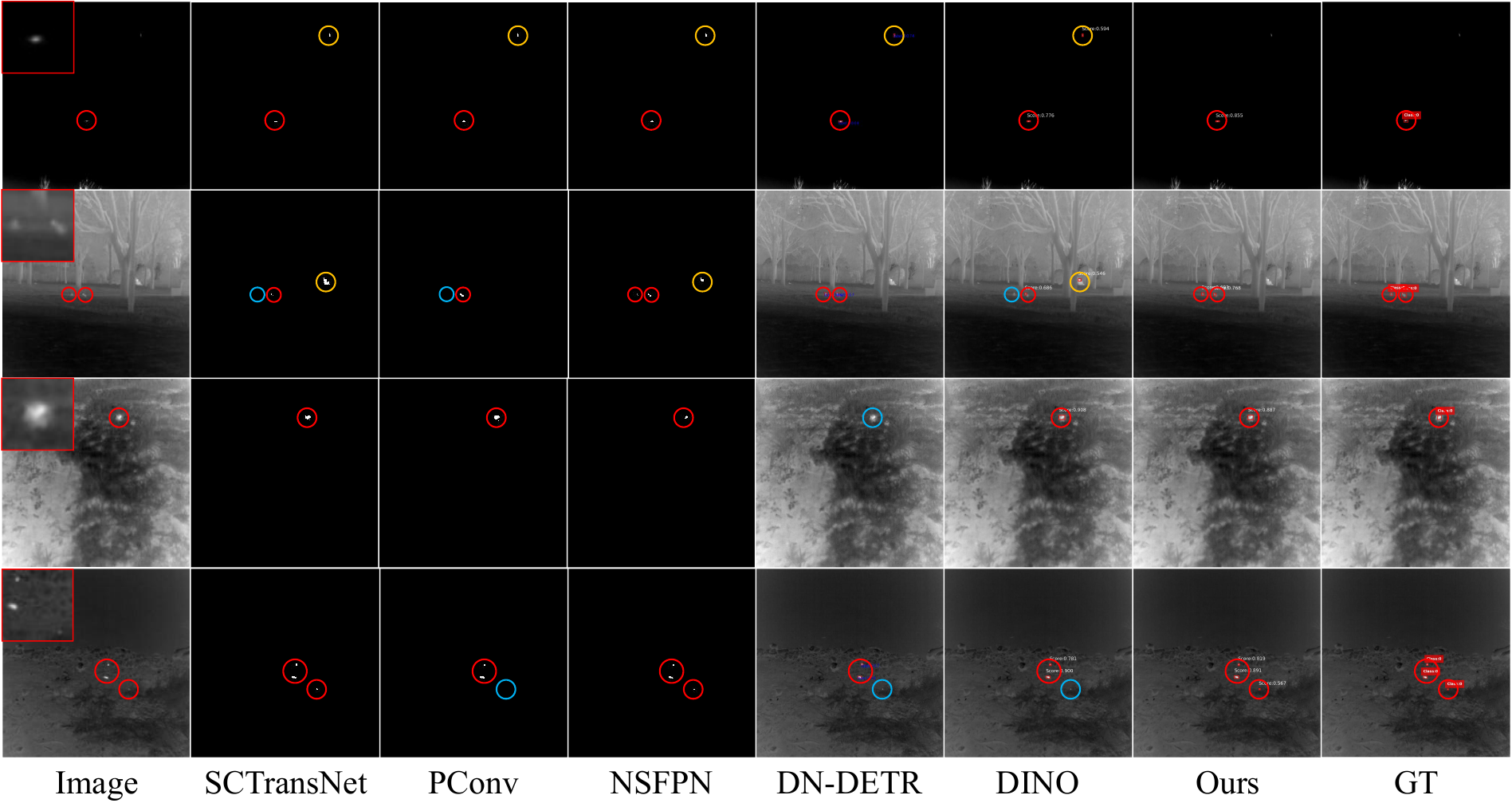}
   \caption{Visualization comparison of detection results via different methods on IRSTD-1k datasets, indicate the land, forests and skies interfere. The red, yellow, and blue boxes denote correct detection, false alarms, and missed detections, respectively.}
   \label{fig:qualitative}
\end{figure*}

We further compare SEF-DETR with representative segmentation-based IRSTD methods on three datasets. To evaluate them under the detection setting, each connected component in the predicted segmentation mask is converted into its axis-aligned bounding box. Since these converted boxes do not carry confidence scores, AP curves cannot be computed, and only P/R/F1 are reported in Table~\ref{tab:segmentation_compare}. SEF-DETR achieves the best performance on all three datasets, reaching 92.4\%/85.9\%/89.0\% on IRSTD-1k, 94.8\%/97.3\%/96.1\% on NUAA-SIRST, and 100.0\%/96.3\%/98.1\% on NUDT-SIRST.
Figure~\ref{fig:qualitative} further illustrates the performance of other superior detection methods under low contrast, structured clutter, and extremely small target sizes on IRSTD-1k dataset. Our SEF-DETR suppresses false alarms in rows 1--2 and recovers dim targets missed by other superior methods in rows 2--4.

\subsection{Ablation Studies}

\textbf{Ablation on Each Component.} Table~\ref{tab:component_ablation} evaluates SEF components on IRSTD-1k using DINO with a ResNet-50 backbone. PSS combined with either FRE or RCF improves the baseline, while the complete PSS+FRE+RCF model achieves the best result, indicating that encoder examination and query re-ranking provide complementary benefits.

\begin{table}[t]
\centering
\small
\setlength{\tabcolsep}{0.9mm}
\begin{tabular}{lcc|ccccc}
\toprule
\textbf{PSS} & \textbf{FRE} & \textbf{RCF} & \textbf{AP} & \textbf{AP\textsubscript{50}} & \textbf{AP\textsubscript{75}} & \textbf{\#Params(M)} & \textbf{GFLOPs} \\
\hline
 & & & 37.1 & 84.5 & 24.3 & 45.14 & 80.94 \\
\checkmark & \checkmark & & 38.3 & 85.0 & 27.1 & +1.56 & +0.22 \\
\checkmark & & \checkmark & 38.1 & 85.7& 26.9 & +0.06 & +0.17 \\ 

\checkmark & \checkmark & \checkmark & \textbf{38.9} & \textbf{86.7} & \textbf{27.1} & +1.56 & +0.22 \\
\bottomrule
\end{tabular}
\caption{Component-wise ablation on IRSTD-1k dataset.}
\label{tab:component_ablation}
\end{table}

\noindent \textbf{Ablation of the patch size $p$ in PSS.} We evaluate different patch sizes $p \in {32,64,96,128}$ in PSS module. As shown in Table~\ref{tab:patch_size}, $p=64$ achieves the best results across all metrics. Smaller patches provide insufficient contextual information, whereas larger patches introduce excessive background interference. Therefore, we set $p=64$ in PSS module.


\begin{table}[t]
\centering
\small
\setlength{\tabcolsep}{2mm}
\renewcommand{\arraystretch}{1.0}
\begin{tabular}{l|ccc|ccc}
\toprule
\textbf{Patch-size $p$} & \textbf{P} & \textbf{R} & \textbf{F1} & \textbf{AP} & \textbf{AP\textsubscript{50}} & \textbf{AP\textsubscript{vt}} \\
\midrule
32 & 89.9 & 83.2 & 86.4 & 38.1 & 84. & 31.3 \\
\textbf{64} & \textbf{92.4} & \textbf{85.9} & \textbf{89.0} & \textbf{38.9} & \textbf{86.7} & \textbf{32.8} \\
96 & 88.2 & 84.9 & 86.5 & 37.6 & 85.7 & 30.1 \\
128 & 89.6 & 83.9 & 86.7 & 38.0 & 85.1 & 30.3 \\
\bottomrule
\end{tabular}
\caption{Ablation study on patch size $p$ in PSS module.}
\label{tab:patch_size}
\end{table}

\noindent \textbf{Different fusion strategies in RCF.} Table~\ref{tab:rcqs_fusion} compares direct addition with reliability-only, consistency-only, and full RCF. Reliability discounts uncertain frequency predictions, while consistency rewards agreement between spatial and frequency scores. Their combination achieves the best result, supporting their complementary roles in query re-ranking.

\subsection{Model Complexity Analysis}
As shown in Tables~\ref{tab:detection_compare}and~\ref{tab:component_ablation}, the complete SEF design adds 1.56M parameters and 0.22 GFLOPs to DINO. When integrated into RT-DETR and LW-DETR, it introduces only 0.31M parameters and 0.18 GFLOPs, while improving AP by 1.2 and 2.9 points, respectively. Most additional parameters arise from the layer-wise projections in FRE, whereas PSS accounts for most of the added frequency-processing computation and RCF performs lightweight score re-ranking with negligible overhead.

\begin{table}[t]
\centering
\small
\setlength{\tabcolsep}{2.0mm}
\renewcommand{\arraystretch}{1.0}
\begin{tabular}{l|ccc|ccc}
\toprule
\textbf{Fusion Factors} & \textbf{P} & \textbf{R} & \textbf{F1} & \textbf{AP} & \textbf{AP\textsubscript{50}} & \textbf{AP\textsubscript{vt}} \\
\midrule
Simply addition & 89.5 & 82.6 & 85.9 & 37.6 & 85.4 & 31.2 \\
Reliability (R) & 88.6 & 83.6 & 86.0 & 37.9 & 85.9 & 30.1 \\
Consistency (C) & 88.0 & 83.9 & 85.9 & 37.4 & 86.1 & 30.4 \\ 
\textbf{R + C (Ours)} & \textbf{92.4} & \textbf{85.9} & \textbf{89.0} & \textbf{38.9} & \textbf{86.7} & \textbf{32.8} \\
\bottomrule
\end{tabular}
\caption{Ablation of different fusion factors in RCF module.}
\label{tab:rcqs_fusion}
\end{table}

\section{Conclusion}

In our paper, we revisited self-attention from the embedding dilution perspective and revealed that the target-relevant embeddings of IRST are inevitably overwhelmed by dominant background features. To address this issue, we observe that the Fourier spectrum of local patches provides discriminative cues for infrared small targets. Building upon this key insight, we proposed SEF-DETR, a pioneering framework that significantly improves target-relevant embedding quality and query initialization by introducing frequency-domain priors through our Patch-wise Spectral Screening, Frequency-Routed Examination, and Reliability-Consistency-aware Fusion. Extensive experiments on the three public IRSTD datasets demonstrate that our SEF-DETR achieves superior performance over previous state-of-the-art object detectors, while introducing only marginal computational overhead. Moreover, the proposed modules can serve as plug-and-play components that are easily transferable to various DETR-like frameworks, providing an effective DETR-based detector for the infrared small target detection task.

\bibliography{main}
\end{document}